\documentclass[10pt,twocolumn,letterpaper]{article}

\usepackage{cvpr}              

\usepackage{graphicx}
\usepackage{amsmath}
\usepackage{amssymb}
\usepackage{booktabs}
\usepackage{pifont}
\usepackage[accsupp]{axessibility}

\usepackage[pagebackref,breaklinks,colorlinks]{hyperref}

\usepackage[capitalize]{cleveref}
\crefname{section}{Sec.}{Secs.}
\Crefname{section}{Section}{Sections}
\Crefname{table}{Table}{Tables}
\crefname{table}{Tab.}{Tabs.}
\crefname{equation}{Eq.}{Eqs.}

\def\cvprPaperID{32} 
\def\confName{CVPR}
\def\confYear{2023}

\begin{document}

\title{Transformer-based Fusion of 2D-pose and Spatio-temporal Embeddings for Distracted Driver Action Recognition}

\author{Erkut Akdag\textsuperscript{\dag}, Zeqi Zhu, Egor Bondarev, Peter H.N. de With\\
VCA Group, Department of Electrical Engineering, Eindhoven University of Technology\\
P.O. Box 513, Eindhoven 5612AZ, The Netherlands\\
\small{\textsuperscript{\dag}~\textit{corresponding author}}\\
{\tt\small \{e.akdag, z.zhu, e.bondarev, p.h.n.de.with\}@tue.nl}\\
}
\maketitle

\begin{abstract} 
Classification and localization of driving actions over time is important for advanced driver-assistance systems and naturalistic driving studies. Temporal localization is challenging because it requires robustness, reliability, and accuracy. In this study, we aim to improve the temporal localization and classification accuracy performance by adapting video action recognition and 2D human-pose estimation networks to one model. Therefore, we design a transformer-based fusion architecture to effectively combine 2D-pose features and spatio-temporal features. The model uses 2D-pose features as the positional embedding of the transformer architecture and spatio-temporal features as the main input to the encoder of the transformer. The proposed solution is generic and independent of the camera numbers and positions, giving frame-based class probabilities as output. Finally, the post-processing step combines information from different camera views to obtain final predictions and eliminate false positives. The model performs well on the A2 test set of the 2023 NVIDIA AI City Challenge for naturalistic driving action recognition, achieving the overlap score of the organizer-defined distracted driver behaviour metric of 0.5079.
\end{abstract}

\section{Introduction} 
\label{sec:intro}
In recent years, the advanced driving-assistance systems~(ADAS) development has garnered significant attention for its promise to improve road safety and reduce accidents. However, accurately identifying and locating hazardous events, particularly in complex driving scenarios, remains a challenge for ADAS development. The recognition of distracted driver activity is an important objective with practical relevance in this area, and it is a crucial cause for fatalities in traffic accidents. For instance, distracted driver activities cause nine fatalities in the United States daily, based on National Highway Traffic Safety Administration statistics~\cite{NationalHighway2021}. Therefore, identifying and addressing such driver behaviour is vital for ADAS and can potentially reduce deaths in traffic incidents.

With recent deep learning advances, the research on modelling systems is increasingly based on examining a substantial volume of video data. One important analysis domain is detection of actions in untrimmed or extended videos, which enables recognizing distracted driver behavior and avoiding accidents. In the last decade, immense research has been conducted, especially on advanced video action recognition techniques to facilitate robust action classification. Nevertheless, this problem still needs further work to be solved due to its challenging nature. For instance, complex motion patterns, actions with similar visual appearances, and varying duration make this problem quite challenging. 

\begin{figure}[t]
  \centering
  \includegraphics[width=\columnwidth]{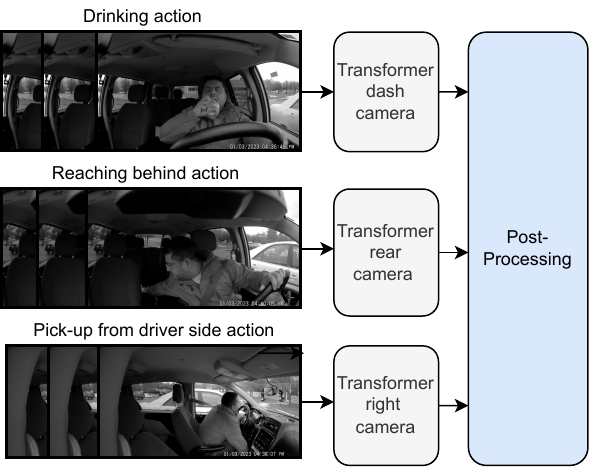} 
  \caption{Overview of the proposed transformer-based method. From left top to bottom: example images of the "drinking", "reaching behind", and "pick-up from driver side" action classes.
  }
  \vspace{-0.7cm}
  \label{fig:intro}
\end{figure}
Distracted driver actions take place in different directions in the space surrounding the driver. To understand these actions for reliable recognition, several cameras are used in different positions. The actions captured by the individual camera views are learned, and then combined~\cref{fig:intro}. In this paper, we propose a transformer-based fusion of 2D-pose and spatio-temporal features to efficiently predict the temporal localization of distracted driver activity. Specifically, we first extract 2D-pose and spatio-temporal features from each video. Secondly, to take advantage of multi-modality of the feature data, we apply a transformer-based model for the feature fusion. 2D-Pose features are given as a position embedding at the encoder stage, named "POSEition" embedding throughout the paper, while spatio-temporal features are the main input to the encoder. During the training step, we use the density-guided label-smoothing technique proposed earlier in~\cite{alkanat2022density}. In the final step, the data from transformer models which are trained on different camera views, is combined and a post-processing step is applied for obtaining temporal localization of distracted driver actions. The method is independent of camera numbers inside the vehicle, making it a generic solution. Summarizing, this paper provides the following contributions

\begin{itemize}
    \item Novel solution for the distracted driver action recognition, based on a transformer model that is independent of the amount of in-vehicle cameras.
    \item Efficient feature extraction from 2D-pose estimation including the key points of the face and hand.
    \item Fusion of 2D-pose features and video action features by the encoder module with multi-head attention. 
\end{itemize}

The remainder of the paper is organized as follows. \Cref{sec:rw} reviews related work on 2D-pose estimation and human activity recognition. In~\Cref{sec:method}, we describe the proposed methodology in detail. \Cref{sec:exp} presents the experimental settings and obtained results, including the ablation studies. Finally, concluding remarks are given in~\Cref{sec:conc}.
\section{Related Work} 
\label{sec:rw}
We review the closely related literature on human activity recognition from three perspectives: video action recognition, distracted driver action recognition, and modalities for feature representation.

\textbf{Video action recognition} 
Action recognition has been the subject of significant research in recent years, with various approaches being used~\cite{tran2018closer,zhao2018trajectory,yang2020temporal,lin2019tsm, feichtenhofer2020x3d}. This field has progressed with modern benchmarks on the Kinetics dataset, offering a larger number of categories and videos. Over the past two decades, the trend in action recognition has been to gather larger datasets and develop larger models (such as C3D~\cite{tran2015learning}, I3D~\cite{carreira2017quo}, and SlowFast~\cite{feichtenhofer2019slowfast}). In addition to this, several backbone architectures, such as VGG~\cite{simonyan2014very}, ResNet~\cite{resnet}, and DenseNet~\cite{huang2017densely}, have been developed to enhance precision and efficiency. While CNNs remain the primary models for this task, Vision Transformers are gradually gaining prominence due to their high potential. The ViT model (Vision Transformer)~\cite{dosovitskiy2020image} applies the transformer architecture directly to image classification and has shown promising results. In recent years, ViT and its variants, including MViT~\cite{fan2021multiscale} and Swin Transformer~\cite{liu2022swin}, have achieved outstanding performance in action recognition.

\textbf{Distracted driver action recognition} 
Distracted driving is a severe problem that has received increasing attention in recent years. Researchers have investigated various approaches to mitigate the effects of distracted driving. The work in~\cite{eraqi2019driver} introduces the distracted driver dataset~(DDD) involving 44~drivers. Authors train multiple convolutional neural network~(CNN) architectures and use a learnable weighted ensemble technique for real-time driver distraction identification. Another work~\cite{martin2019drive} publishes the Drive\&Act dataset, including 12 distraction-related actions captured by six cameras. The authors show that their video and body pose-based action recognition algorithm gives competitive results. In the study by~\cite{kopuklu2021driver}, the authors propose a contrastive learning approach to learn a metric to differentiate normal driving from abnormal driving. To achieve this, they introduce the Driver Anomaly Detection~(DAD) dataset, which consists of 31~subjects that perform various activities in a real car.

\begin{figure*}[ht]
  \centering
  \includegraphics[width=1\linewidth]{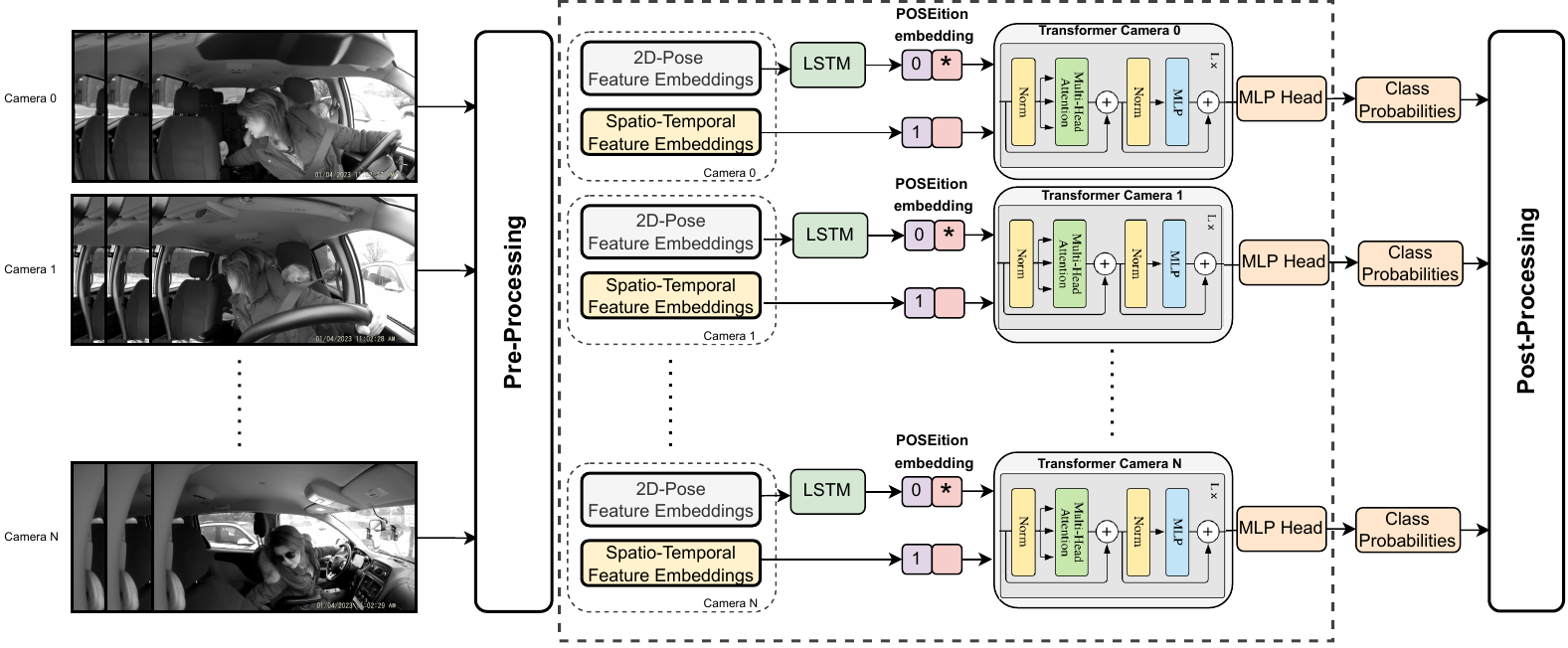} 
  \caption{Overview of the proposed architecture. Left: different camera-view inputs to the pre-processing step. Middle: extracted 2D-pose and spatio-temporal embeddings are supplied to the transformer architecture. 2D-Pose embedding is considered as the "POSEition" embedding of the encoder, while the spatio-temporal embedding is the main input of the encoder. Right: 1D-class probabilities obtained after the MLP head per camera view are analyzed for finding the significant peaks for each class in a video. Note that the modules shown with a bold dashed border are used for training only.}
  \label{fig:framework}
\end{figure*}

\textbf{Modalities for feature representation} 
In previous studies, various feature modalities have been explored and integrated to enhance the performance of video action recognition. An RGB frame, as an atomic video component, is the most common input modality~\cite{inception, resnet, rgb_in_1} for image classification. Ng et al.~\cite{rgb_in_3} extracted features independently from each RGB frame using 2D-ConvNet, followed by LSTM cells~\cite{lstm} for a sequence of ConvNet outputs, to encode state and capture temporal ordering and long-range dependencies. Du et al.~\cite{c3d} have used segmented RGB volumes from the video as inputs for 3D-ConvNet to classify actions, as temporal information is propagated across all the layers in the network, allowing 3D-ConvNet to model temporal information better than 2D-ConvNet. In addition to RGB frames, Simonyan and Zisserman~\cite{optical-flow-2-streams} have introduced optical flow to explicitly describe the human motion between video frames. Complementary modalities of motion significantly boost model performance in action recognition. Recently, the human skeleton data has received increasing attention for its action-oriented nature and compactness. Most works~\cite{DBLP:journals/corr/abs-1907-13025, potion, mmtm, co-o, multitask, two_stream, skeleton-base-method} extract body skeletons from raw images via 2D-CNN-based approaches and convert the coordinates in a skeleton sequence to a pseudo-image with transformations. The skeleton features can be fused with other modalities and further processed by graph convolutional network~(GCN)-based~\cite{ST-GCN} or 3D-CNN-based approaches~\cite{two_stream, skeleton-base-method}, achieving superior performance on action recognition benchmarks. 

Previous works have demonstrated that representations from different modalities, such as RGB and skeletons, are complementary. In this work, we propose to integrate  the pose features of the driver face and body parts with spatio-temporal video features to enhance performance in action recognition.
\section{Methodology} 
\label{sec:method}
We propose a transformer-based end-to-end fusion solution to exploit spatio-temporal features aligned with 2D-pose features, as shown in~\Cref{fig:framework}. In a pre-processing step, we re-size each video frame to 512$\times$512~pixels and feed our model to extract 2D-pose features and the spatio-temporal features. The proposed architecture takes 2D-pose features as position embedding, named "POSEition" and spatio-temporal features as the main input to the encoder of the transformer-based fusion model. In the final step, post-processing of the output probabilities of frames in each video identifies significant peaks, related to each class in the corresponding video. 

\subsection{2D-Pose Features}
\label{sec:2d_pose_extraction}
\begin{figure}[b]
\centering
\includegraphics[scale=0.25]{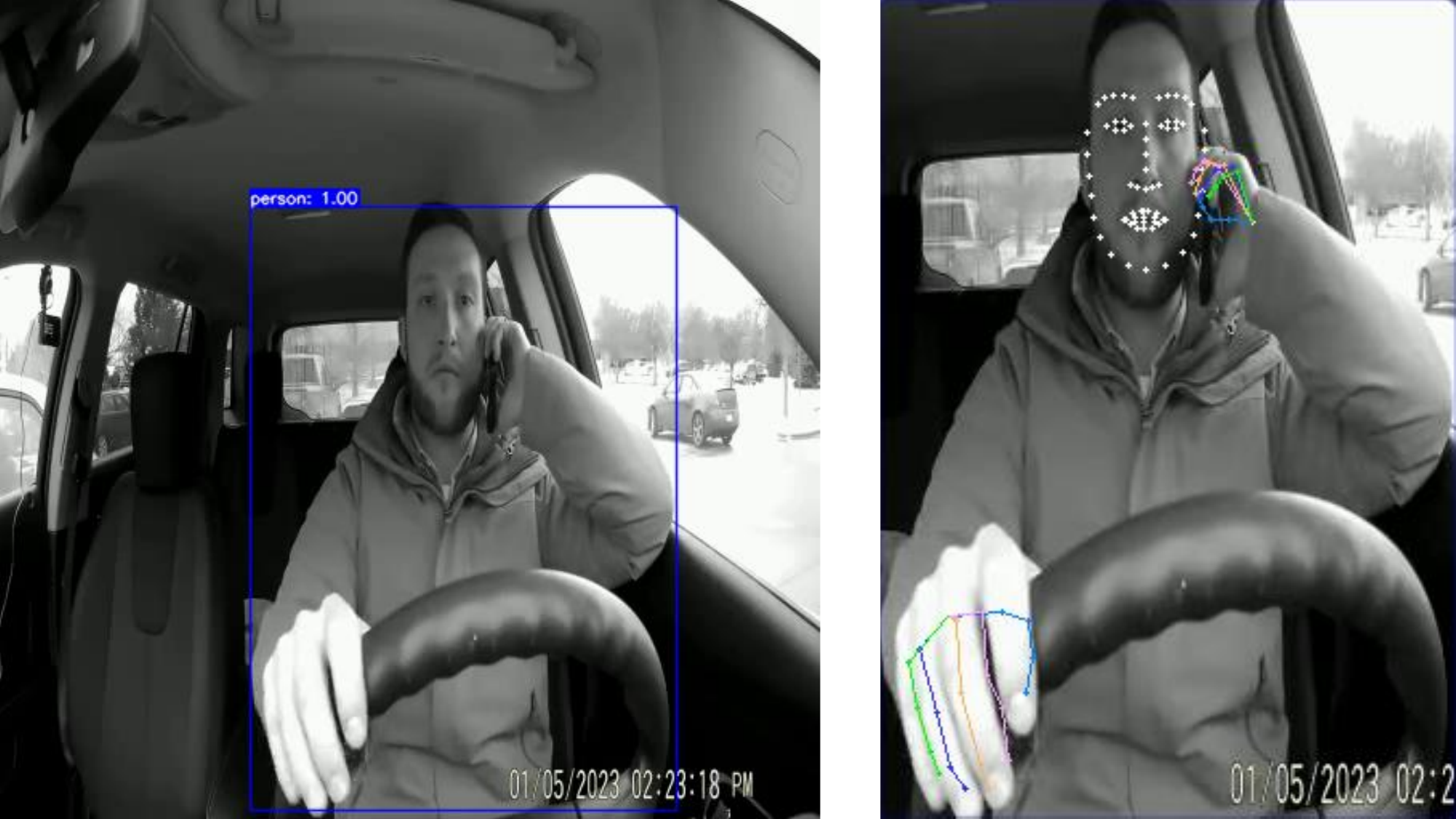}
\caption{Top-Down pose estimation.}
\label{fig:top_down_pose_estimator}
\end{figure}

Previous studies~\cite{skeleton-base-method, gcn-method} have highlighted the critical importance of 2D-pose extraction in the pre-processing step of skeleton-based action recognition. Even small perturbations in the joint coordinates can lead to vastly different predictions. To ensure accurate pose extraction, we adopt a 2D Top-Down pose estimator~\cite{hrnet}, which has been shown to outperform its 2D Bottom-Up counterparts on standard benchmarks~\cite{coco}. As described in~\Cref{fig:top_down_pose_estimator}, we first utilize a Faster R-CNN~\cite{DBLP:journals/corr/RenHG015} human detector which is pre-trained on the COCO dataset, to identify the region of the driver in the raw RGB frame. Next, we input the detected region to the Top-Down pose estimator HRNet~\cite{hrnet}, which generates the whole-body joint coordinates of the driver. Since the focus is solely on a single driver, the Top-Down pose estimator offers higher accuracy than the Bottom-Up estimator, while requiring the same processing time.

\begin{figure}[b]
     \centering
     \begin{subfigure}[b]{0.23\textwidth}
         \centering
         \includegraphics[height=0.22\textheight, width=0.8\textwidth]{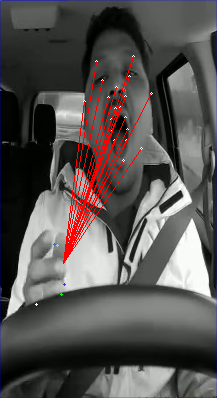}
     \end{subfigure}
     \hfill
     \begin{subfigure}[b]{0.23\textwidth}
         \centering
         \includegraphics[height=0.22\textheight, width=0.8\textwidth]{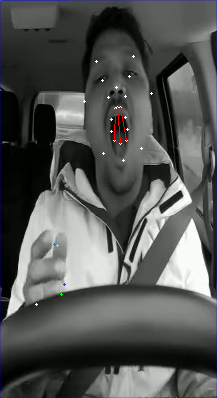}
     \end{subfigure}
        \caption{Relative distances between hand and face points (left), and set of facial feature points (right).}
    \label{fig:distance}
\end{figure}

\begin{figure}[b]
     \centering
     \begin{subfigure}[b]{0.23\textwidth}
         \centering
         \includegraphics[height=0.22\textheight, width=0.8\textwidth]{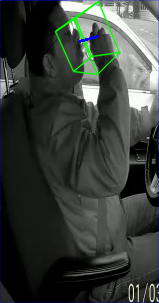}
     \end{subfigure}
     \hfill
     \begin{subfigure}[b]{0.23\textwidth}
         \centering
         \includegraphics[height=0.22\textheight, width=0.8\textwidth]{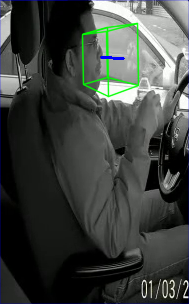}
     \end{subfigure}
        \caption{Head pose of a driver during a drinking action.}
    \label{fig:head_rotation}
\end{figure}

\begin{figure}[t]
     \centering
     \begin{subfigure}[b]{0.23\textwidth}
         \centering
         \includegraphics[height=0.22\textheight, width=0.8\textwidth]{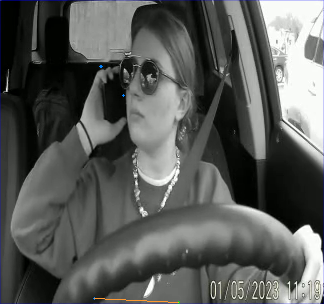}
     \end{subfigure}
     \hfill
     \begin{subfigure}[b]{0.23\textwidth}
         \centering
         \includegraphics[height=0.22\textheight, width=0.8\textwidth]{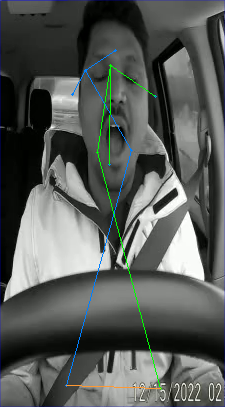}
     \end{subfigure}
        \caption{ Missed (left) and flawed (right) leg estimations for the driver.}
    \label{fig:wrong_predictions_on_legs}
\end{figure}

For the whole-body pose estimation, a total of 133 key points are used (17 for the body, 6 for the feet, 68 for the face, and 42 for the hands)~\cite{coco_keypoints}. However, many of these key points are irrelevant and do not aid in action recognition. For instance, when the driver's body is occluded by the steering wheel (as shown in~\cref{fig:wrong_predictions_on_legs}), the estimation of key points for both legs can be either inaccurately estimated or completely missed, which is likely to interfere with the final prediction results. Therefore, we select a subset of critical key points from body parts that correspond highly to distracted driving activities, such as the face, shoulders and hands. As illustrated in~\cref{tab:cls_related_move}, actions such as drinking, phone calling, eating, texting, reaching behind, and picking-up objects are strongly associated with hand movements, while head pose and facial expression are highly relevant for actions like singing, yawning, talking, drinking, and phone calling.

The driver's skeleton can be represented as a set of joint coordinates, where $J_i = (x_i,y_i,c_i)$ denotes the location and confidence score of a detected joint $i$, $i=1..N_j$. In time, a skeleton sequence of $T_c$ frames can be represented as a $T_c \times N_j \times 3$ array, where $N_j$ is the number of selected joints per skeleton. The joint coordinates alone cannot explicitly convey motion information between body parts to the network. To address this, we focus on highly active body parts (face and hands) and incorporate additional features into the pose feature embeddings. These features include the pose of the head $P^{h}$, the relative positions of the left and right hands to the face $D^{h}$, and the relative position of the upper lip to the lower lip $D^{l}$,  as depicted in~\cref{fig:distance} and \cref{fig:head_rotation}. 

To estimate the driver's head pose, we implement the Perspective-n-Point~(PnP) pose computation method~\cite{pnp}. The resulting rotation and translation vectors are used to draw the orthogonal line $P_h$ of the driver's face, which indicates the viewing direction of the driver. In the visual representation provided in~\cref{fig:head_rotation}, the variation in the length and position of the blue line in accordance with the driver's head movement, provides a clear indication of the driver's head pose.
 
\begin{table}[ht]
\centering
\caption{Classification of driver actions based on the presence of the driver's head, mouth, and hands movements. The \ding{51} symbol indicates the presence of a movement, while an empty cell indicates its absence.}
\label{tab:cls_related_move}
\begin{tabular}{l|c|c|cc}
\toprule
 \textbf{Class} & \textbf{Head} & \textbf{Mouth} & \textbf{Hands} \\
 \midrule
Normal driving &  &  & \ding{51} \\  
Drinking & \ding{51}\ & \ding{51}\ & \ding{51} \\  
Phone call (right hand) &  & \ding{51} & \ding{51} \\ 
Phone call (left hand) &  & \ding{51} & \ding{51} \\ 
Eating                &  &  & \ding{51} \\ \hline
Texting (right hand) & \ding{51} &  & \ding{51} \\ 
Texting (left hand) & \ding{51} & & \ding{51}\\ 
Reaching behind & \ding{51} &  & \ding{51}\\ 
Adjusting control panel & &  & \ding{51} \\ \hline
Picking up (driver) & \ding{51} &  & \ding{51} \\ 
Picking up (passenger)& \ding{51} &  & \ding{51} \\ 
Talking to passenger (right) &  \ding{51} & \ding{51} &  \\ 
Talking to passenger (back)  &  \ding{51} & \ding{51} & \\ \hline
Yawning & & \ding{51} & \ding{51}  \\ 
Hand on head  & \ding{51} &  & \ding{51} \\ 
Singing  & \ding{51} & \ding{51} & \ding{51} \\ 
\bottomrule
\end{tabular}
\end{table}
To capture the relative distance between two body joints $J_i$ and $J_j$ with Cartesian coordinates $(x_i, y_i)$ and $(x_j, y_j)$ in the 2D plane, we calculate the distance between joints $J_i$ and $J_j$ using~\cref{eq:dist}. By considering the relative distance between joints, we can describe the temporal dynamics of the driver's mouth and hands, allowing for a more comprehensive understanding representation of the driver's behaviour in terms of mouth and hands. The distance between the joints $J_i$ and $J_j$ is specified by
\begin{equation}
D(J_i, J_j) = \sqrt{(x_i-x_j)^2+(y_i-y_j)^2}.
\label{eq:dist}
\end{equation}
To incorporate movement information of the above active body parts, we collect the aforementioned pose and distance features at each frame and arrange them as a chain of motion vectors over time $T_c$. The motion vector is described by $fm_{t} = (P^{h}, D^{h}_{0},...,D^{h}_{L}, D^{l}_{0},...,D^{l}_{K})_{t}$, where $t$ denotes the time step, $L$ is the number of relative distances between hand joints and face joints, $K$ is the number of relative distances between upper-lip joints and bottom-lip joints. We then concatenate the selected skeleton coordinates with the motion vector and use the resulting feature embeddings of the $i^{th}$ video, $f\text{-pose}_{t}^i  = (J_0,...,J_{N_j}, P^{h}, D^{h}_{0},...,D^{h}_{L}, D^{l}_{0},...,D^{l}_{K})_{t}^i$, as the additional input stream for the proposed network. This allows the model to leverage both the spatial and temporal dynamics of the driver's behavior for improving the prediction accuracy.
\subsection{Spatio-temporal Features}
To extract spatio-temporal features, we employ the SlowFast video action recognition network pre-trained on the Kinetics-400 dataset~\cite{kay2017kinetics}. This network has a multi-branch architecture that captures both short-term and long-term relations, by extracting one low-rate input sequence (fast pathway) and one high-rate input sequence (slow pathway) through temporal sub-sampling of the video frames. The fast pathway helps to describe long-term actions, while the slow pathway is better suited for short-term actions. Considering multiple temporal resolutions enables to extract features robustly for both types of actions. The SlowFast architecture takes every video segment of the provided training videos as input. Formally, assume that $V_s^i\in\mathbb{R}^{h\times{}w\times{}T_c}$ denotes a video segment that consists of $T_c$ frames at the time interval $[t, t+T_c-1]$ of the $i^{th}$ video with horizontal and vertical resolution of $h$ and $w$, respectively. The output spatio-temporal feature vector, $f\text{-spatem}_t^i = S(V_s^i)\in\mathbb{R}^{N_f}$, is used for further capturing spatio-temporal embeddings for the related video segment, where $S$ is the functional form of the SlowFast backbone and $N_f$ is the resulting feature size. In the proposed solution, SlowFast extracts spatio-temporal embeddings for each 64-frame segment of the video clip with a temporal stride of unity. The size of the output feature vector, $N_f$, is 2,304 for each video segment.
\subsection{Fusion of 2D-pose and Spatio-temporal Embeddings}
Feature concatenation and ensembling have been widely used in various machine learning applications, and their effectiveness have been demonstrated in numerous studies. Feature concatenation involves combining multiple features or variables into a single input, which can help capturing complex relationships between them. Although feature concatenation and ensembling can be efficient techniques for improving the performance of machine learning models, there are also some potential drawbacks that should be considered. For instance, if the concatenated features are not carefully selected or pre-processed, this can increase data noise which can harm the model's performance. These drawbacks can make it more challenging to interpret the contributions of individual features to the model's predictions. Furthermore, if the data modalities used for ensembling are too similar or have similar weaknesses, this may not improve performance. For instance, ensembling only spatio-temporal models or 2D-pose-based models may not improve the results as expected. Therefore, we propose an efficient transformer-based architecture to fuse the 2D-pose features and spatio-temporal features to address the aforementioned drawbacks. By using this fusion approach, we improve the robustness of the model, and obtain a more accurate overall prediction and reduce the risk of over-fitting. 

The dashed box in the middle of the architecture in~\Cref{fig:framework} illustrates the transformer-based model for fusing the 2D-pose and spatio-temporal embeddings. In previous sections, we have obtained the 2D-pose output feature vector $f\text{-pose}_{t}^i = (J_0,...,J_{N_j}, P^{h}, D^{h}_{0},...,D^{h}_{L}, D^{l}_{0},...,D^{l}_{K})_{t}^i$ and the spatio-temporal output feature vector $f\text{-st}_t^i$. The proposed architecture includes one LSTM layer between the 2D-pose output feature vector and the POSEition embedding input of the encoder for two main reasons. Firstly, the POSEition embedding dimensionality should be consistent with the main encoder input size of spatio-temporal embedding. Secondly, the LSTM layer brings additional temporal information from the extracted 2D-pose features. Therefore, the LSTM module takes $f\text{-pose}_{t}^i$ as input and provides the output feature vector size $N_f$=2,304 for the POSEition embedding. While the primary input of the encoder module part of the transformer-based fusion model is spatio-temporal features, 2D-pose feature embeddings become the POSEition embedding input. The transformer encoder consists of the multi-head self-attention (MSA) with normalization and MLP blocks. The MLP contains two layers with a GELU activation $GELU(x) = x\varphi(x)$, where $\varphi(x)$ is the cumulative distribution function of the standard Gaussian distribution. Layer normalization (LN) is applied prior to each block and residuals are concatenated after each block. The output of the transformer encoder is fed into the MLP head for computing the class probabilities per camera view.

\begin{figure}[t]
  \centering
  \includegraphics[width=\columnwidth]{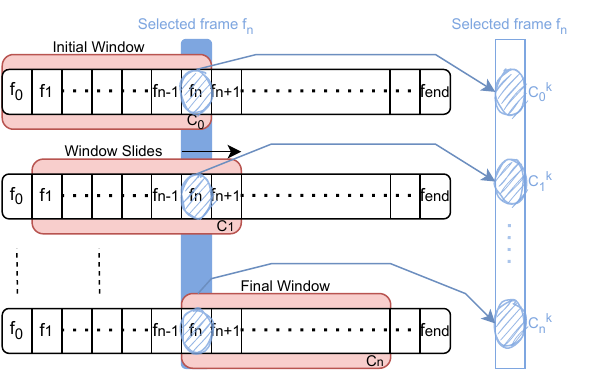} 
  \caption{Illustration of one selected frame based on the sliding window approach.
  }
  \vspace{-0.7cm}
  \label{fig:intro4}
\end{figure}

The proposed method uses a temporal sliding-window approach during training phase to obtain features for each video frame. In this problem, sliding window can be defined as a video segment and the size of the sliding window is equal to the number of frames in that video segment. After obtaining the features for each frame, method combines and utilizes them during the training. As stated in ~\Cref{fig:intro4}, assuming the $C_n^k$ denotes a video segment with a segment size $k$ that consists of frame-based class probabilities of $n^{th}$ frame of a video after applying the sliding window technique. Each frame has $k$ different class probability values from the sliding window technique, where the $f_n$ refers to the selected $n^{th}$ frame within a video segment size $k$. In the proposed architecture, video segment size is set $k$=64. Considering the selected frame $f_n$ the label with the maximum number of frames in a given video segment is regarded as the label for the entire video segment for the batch sampling. However, there is a problem with using multi-label video segments during training, making it difficult for the network to classify them as single labels. Typically, researchers deal with this by either using the most commonly observed frame-level label as the segment-level label during training or by discarding these segments altogether. However, authors in~\cite{alkanat2022density} show the usefulness of the density-guided label smoothing technique for improving the network's overall performance by re-using information located within boundary segments. Based on this approach, we train our network with the cross-entropy loss for classification. Including the density-guided label smoothing, the complete loss function is given as
\begin{equation}
    \mathcal{L}(x) = -\sum_{k}p_t^i(k)q^{\prime\prime}(k|x),
\end{equation}
where $(i, t)\in B$ defines a unique video segment in a training-batch~$B$. We apply the generalized form of the softmax function with the temperature parameter $\beta$, enabling the control over smooth vs. sharp transitions of the class label change. After training, we supply test video segments to our network, extracting segment class probabilities, $p_t^i(y)$, which are further subject to subsequent post-processing, explained in the next section. 

\subsection{Post-Processing}
Multi-camera setup requires the combination of individual stream probabilities into global frame-level probabilities. This means that the class probabilities for a frame from all camera views should be fused to extract the class probabilities of the multi-camera scene. To achieve this, we average all the segment-level class probabilities associated with a video segment that contains the frame under consideration, which is represented by
\begin{equation}
    P_\text{frame}(y) = \frac{\displaystyle\sum_{c=1}^{C}P_\text{frame}^{seg}(y)}{c}.
\end{equation}
where $P_\text{frame}(y)$ is the output frame-level class probabilities within a associated video segment, $P_\text{frame}^{seg}(y)$ is the segment-level class probabilities for that selected frame under consideration, and $c$ denotes the total number of cameras within the multi-camera setup, which is set to $c$=3 in this problem. As a result, the likelihood value assigned to a combination of a class label and a selected scene frame includes information about the likelihood of other class probabilities for that frame, providing 1-D signal output for specified frame including all class probabilities. 

Raw predictions of temporally localized and classified actions can be obtained by detecting consistent peaks in the frame-level scene probabilities for each class and scene separately. To identify consistent peaks in obtained 1-D signal, we first use median temporal filtering to remove noise while maintaining sharp edges that are helpful for accurate action localization. To achieve this, we determine the fastest positive and negative changes that precede and succeed at the highest peak to determine the start and end times of the detected driver behaviour. 
However, some actions may appear similar, resulting in high probabilities for more than one action being assigned to the same frame in individual scenes. Since the target dataset specifies only one action for a given time instance in all scenes, overlapping predictions can increase the false-positive rate and decrease the overall precision of the methodology. To overcome this, we first calculate the intersection-over-union (IoU) for all possible pairs of predictions within the same scene. Then, among the complete set of predictions with an IoU overlap greater than a pre-defined threshold ($o_{max}$), they only keep the prediction with the highest peak in the final output.

\section{Experiments} 
\label{sec:exp}
\subsection{Track3 Dataset}
The 2023 NVIDIA AI City Challenge~\cite{Naphade23AIC23} 
has released naturalistic driving dataset~\cite{Rahman22SynDD2} consisting of 210 video clips, of around 10-min. length each with a resolution of 1920$\times$1080~pixels. Videos recorded of the driver are from various angles using three synchronized cameras mounted inside a car. The dataset features 35~drivers, some wearing appearance blocks such as sunglasses and hats. Each driver performs 16~different tasks such as phone calls, eating, reaching back, in random order, within each video clip. The dataset is split into three parts, A1, A2, and B. The A1 dataset includes labels indicating the start time, end time, and the class of distracting behavior performed, while the A2 dataset has no labels. The main goal of the challenge is to localize and classify the distracted behavior activities, performed by the driver during a given time frame on the A2 test set. The B dataset is reserved for later testing, to determine the top-performers of the challenge.
\subsection{Implementation Details}
For training, we use four GTX-1080Ti GPUs and set the batch size to 80. This batch size allows incorporating 5~training samples of each action class. We adopt the Adam optimizer with initial learning rate 5$\times{10}^{-6}$ and weight decay 5$\times{10}^{-4}$. The proposed architecture uses the cross-entropy loss with the density-guided label smoothing as the loss function. The temperature parameter, $\beta$, is set to $\beta$=5. The size of the resulting output feature vector of the 2D-pose feature extraction step is 460. One layer LSTM takes this and provides a 2,304~dimensional vector as output into the POSEition embedding. The resulting output spatio-temporal feature vector has a size of~2,304, which is compatible with the LSTM output layer. We apply a temporal median filter with a filter size of 351~frames for temporal localization and post-processing step. We set the minimum height and width to~0.1 and 200~frames, respectively, to detect peaks, based on empirical observations.

\subsection{Evaluation metrics}
The evaluation metric of the 2023 NVIDIA AICITY challenge Track3 is the overlap score. Given a ground-truth class activity $g$ with start time $gs$ and end time $ge$, and a predicted class activity $p$ with start time $ps$ and end time $pe$, with the added condition that the start time $ps$ and the end time $pe$ are in the range $[gs-10s,gs+10s]$ and $[ge-10s,ge+10s]$, respectively. To find the overlap between $g$ and $p$, the overlap score is defined as the ratio between the time intersection and the time union of the two activities, i.e., 
\begin{equation}
    os(p,g) = \frac{\max(\min(pe,ge)-\max(gs,ps),0)}{\max(ge,pe)-\min(gs,ps)},
\end{equation}
where the final score is the average overlap score among all matched and unmatched activities. In addition to the overlap score, we include the $F_1$ score, precision, and recall metrics to evaluate more metrics in the experimental results, based on the changes performed in different experiments. The $F_1$ score is specified by
\begin{equation}
  F_1 = 2 \times \frac{P_r\cdot R_c}{P_r+R_c}.
  \label{eq:f1score}
\end{equation}

The precision $P_r$ is calculated as the number of true positives $N_\text{TP}$ divided by the total number of true positives $N_\text{TP}$ and false positives $N_\text{FP}$, which is given by:
\begin{equation}
  P_r = \frac{N_\text{TP}}{N_\text{TP}+N_\text{FP}}.
  \label{eq:precision}
\end{equation}
The recall $R_c$ is calculated as the number of true positives $N_\text{TP}$ divided by the total number of true positives $N_\text{TP}$ and false negatives $N_\text{FN}$, which is given by:
\begin{equation}
  R_c = \frac{N_\text{TP}}{N_\text{TP}+N_\text{FN}}.
  \label{eq:recall}
\end{equation}

\begin{table}[b!]
\caption{Experimental results for the proposed solution with two different settings.}
\resizebox{\columnwidth}{!}{%
\begin{tabular}{l|lccc}
\toprule
Method & $os$ score & $F_1$ score & Precision & Recall \\ \midrule
\begin{tabular}[c]{@{}l@{}}Solution with\\ 2D-Pose skeleton\end{tabular} & 0.4929 & 0.6359 & 0.6591 & 0.5708 \\ \hline
\begin{tabular}[c]{@{}l@{}}Solution 2D-Pose \\ skeleton\&motion\end{tabular} & 0.5079 & 0.6452 & 0.6789 & 0.5783 \\ \bottomrule
\end{tabular}%
}
\label{tab:results}
\end{table}
\begin{table}[b!]
\centering
\caption{Experimental results of different video action recognition models for the ``dash camera view" dataset.}
\begin{tabular}{@{}l|cccc@{}}
\toprule
Method   & $os$ score & $F_1$ score & Precision & Recall \\ \midrule
SlowFast~\cite{feichtenhofer2019slowfast} & 0.3459      & 0.4705      & 0.4965    & 0.4472 \\
X3D~\cite{feichtenhofer2020x3d} & 0.3136      & 0.4367      & 0.4689    & 0.4379 \\
Fine-tuned X3D  & 0.3203      & 0.4399      & 0.4796    & 0.4361 \\ \bottomrule
\end{tabular}
  \label{tab:ablation_video}
\end{table}

\begin{table}[t!]
\centering
\caption{Experimental results of ablation studies.}
\resizebox{\columnwidth}{!}{%
\begin{tabular}{l|lccc}
\toprule
Method & $os$ score & $F_1$ score & Precision & Recall \\ \midrule
Baseline spatio-temp. & 0.3703 & 0.5126 & 0.6120 & 0.4410 \\ \midrule
\begin{tabular}[c]{@{}l@{}}+ concat feat. vect. \\ 2D-pose skel. \end{tabular} & 0.4274 & 0.5987 & 0.6364 & 0.5652 \\ \midrule
\begin{tabular}[c]{@{}l@{}}+ concat feat. vect. \\ 2D-pose skel.\&mot.\end{tabular} & 0.4420 & 0.6330 & 0.6912 & 0.5838 \\ \midrule
\begin{tabular}[c]{@{}l@{}} Solution w. \\ 2D-pose skel. \end{tabular} & 0.4322 & 0.5859 & 0.6397 & 0.5403 \\ \midrule
\begin{tabular}[c]{@{}l@{}} Solution w. \\ 2D-pose skel.\&mot.\end{tabular} & 0.4493 & 0.6381 & 0.6896 & 0.5846 \\ 
\bottomrule
\end{tabular}%
}
    \label{tab:ablation2d_pose}
\end{table}
\subsection{Experimental Results}
We have obtained an $os$ score of~0.5079 in the 2023 NVIDIA Challenge Track3. Note that we have created our local evaluation system during the competition to verify the model's effectiveness and extend our experiments accordingly. The local evaluation system scores have been almost identical to the public leaderboard scores. In the public leaderboard, we have achieved an $os$ score of~0.4849 but with the extended experiments and ablation studies, the proposed method achieved an $os$ score of~0.5079. We present some of our results in \cref{tab:results}, which demonstrates the effectiveness of the proposed approach in evaluating different metrics. This table provides two different 2D-pose features, while one of them includes only skeleton-based points, resulting in a 212-dimensional 2D-pose feature vector. The other feature vector consists of motion vectors additionally based on the selected skeleton points, resulting in a 460-dimensional 2D-pose feature vector. The proposed solution improves the performance by using the 2D-pose features, including both skeleton-based and motion vector-based key points from the selected skeleton points.


\subsection{Ablation Studies}
\label{sec:abs}
We conduct most of the ablation experiments for the ``dash camera view" training dataset to understand the importance of each architecture component. This dataset can provide sufficient information for most of the actions. Throughout the ablation study, the following questions are addressed: 

\noindent\textit{1. Does another video action recognition model work better compared to the SlowFast network?}~We have employed the X3D~\cite{feichtenhofer2020x3d} video action recognition backbone network to compare to the SlowFast results. Both X3D and SlowFast feature extractors are models pre-trained on the Kinetics-400 dataset. \cref{tab:ablation_video} illustrates that the X3D model does not improve the results. The 2D-pose features have not been integrated in this comparison.

\noindent\textit{2. Does fine-tuning on the video action recognition model improve performance?}~We have examined fine-tuning of the X3D feature extractor model on the ``dash camera view" dataset by using transfer learning. However, it does not improve the score. A possible reason for this can be data shortage as shown in~\cref{tab:ablation_video}. Note that we have prepared the naturalistic driver dataset in the format of the Kinetics-400 dataset for this experiment. 

\noindent\textit{3. Does standard vector concatenation of 2D-pose and spatio-temporal features improve performance?}~We have started conducting the experiments with standard feature vector concatenation of the extracted 2D-pose features, based on the coordinates and spatio-temporal features without having the transformer-based architecture. In these experimental settings, we have taken the mean of each 64~chunks from the 2D-pose feature vector and obtained one feature vector per frame. This improves the contribution compared to using only spatio-temporal features, as shown in~\cref{tab:ablation2d_pose}. 

\noindent\textit{4. Which 2D-pose features are better?}~We have applied two different types of 2D-pose features. The first type contains skeleton-joint coordinates $f\text{-pose}_{t}^i  = (J_0,...,J_{N_j})_{t}^i$, resulting in a 212-dimensional vector. The second type includes motion vectors based on the selected skeleton coordinates $f\text{-pose}_{t}^i  = (J_0,...,J_{N_j}, P^{h}, D^{h}_{0},...,D^{h}_{L},D^{l}_{0},...,D^{l}_{K})_{t}^i$, resulting in a 460-dimensional vector. \cref{tab:ablation2d_pose} indicates the performance improvement of the motion-vector model.

\section{Conclusion} 
\label{sec:conc}
In this paper, we have proposed a transformer-based solution for the fusion of 2D-pose and spatio-temporal features to classify and localize the distracted driver behaviors in the time domain. First, we have created both 2D-pose features and spatio-temporal features similar in dimension with the help of a LSTM module. Next, 2D-pose embeddings and spatio-temporal embeddings are fused within the encoder, including the MHA to obtain the frame-level class probabilities for the different camera views. Finally, in the post-processing step, all frame-level probabilities per camera are combined and the final prediction for each distracted driver class is obtained. The proposed solution has been evaluated on the A2 test set of the 2023 NVIDIA AI City Challenge's Track3, yielding a 0.5079 $os$ score.

\section*{{Acknowledgements}}
This work is supported by the European ITEA project SMART on intelligent traffic flow systems.

{\small
\bibliographystyle{ieee_fullname}

}

\end{document}